\pdfoutput=1
\documentclass[11pt]{article}

\usepackage[final]{acl}

\usepackage{times}
\usepackage{latexsym}
\usepackage{multirow}
\usepackage{booktabs}
\usepackage[T1]{fontenc}
\usepackage[utf8]{inputenc}
\usepackage{amsfonts}
\usepackage{amsmath}
\usepackage{amssymb}
\usepackage{microtype}

\usepackage{inconsolata}

\usepackage{graphicx}
\usepackage{blindtext}

\usepackage{geometry}
\usepackage{marvosym}
\usepackage{url}
\usepackage{fancyhdr}
\usepackage{hyperref}

\usepackage{enumitem}
\usepackage{minted}
\usepackage{booktabs}
\usepackage{algorithm}
\usepackage{algorithmic}
\usepackage{amsmath}
\usepackage{subcaption}
\usepackage{pifont}

\usepackage{pgfplots}
\pgfplotsset{compat=1.17}

\usepackage[table,dvipsnames]{xcolor} % For extended color names (like "teal")

\usepackage{colortbl}  
\usepackage{tikz}                    % Loads TikZ and PGF math (needed for \pgfmathparse)

\newcommand{\cval}[2][]{%
  \pgfmathparse{round(100*#2^3)}% Multiply the cube of the number by 100 and round it
  \edef\shadepercent{\pgfmathresult}%
  \edef\mycolor{cyan!\shadepercent}%
  \expandafter\cellcolor\expandafter{\mycolor}#1{#2}%
}

\usepackage{pgf} % for \pgfmath
\newcommand{\ccol}[3]{%
  \begingroup
  \pgfmathsetmacro{\v}{#1}
  \pgfmathsetmacro{\lo}{#2}
  \pgfmathsetmacro{\hi}{#3}
  \pgfmathsetmacro{\den}{max(\hi-\lo,1e-9)}%
  \pgfmathsetmacro{\praw}{(\v-\lo)/\den}%
  \pgfmathsetmacro{\pnl}{max(min(\praw,1),0)}% clamp [0,1]
  \pgfmathsetmacro{\pct}{round(100*\pnl)}%
  \edef\cellcmd{\noexpand\cellcolor{cyan!\pct!white}}%
  \cellcmd \v
  \endgroup
}

\newcommand{\cbert}[1]{\ccol{#1}{0.800}{0.9168}}
\newcommand{\cbleu}[1]{\ccol{#1}{0.40}{69.8534}}

\newcommand{\redx}{\cellcolor{white!40!}\ding{55}} % Using \ding{55} for a cross symbol
\newcommand{\greentick}{\cellcolor{white!90!}\ding{51}} % Using \ding{51} for a tick symbol

\usepackage[most]{tcolorbox}

\tcbset{
  parskip/.style={
    before={\par\pagebreak[0]\parindent=0pt},
    after={\parfillskip=0pt\par},
  },
}
\newtcolorbox{resultbox}[1][]{%
    colback=black!3,
    colframe=black!3,
    notitle,
    sharp corners,
    borderline west={2pt}{0pt}{gray!80!black},
    enhanced,
    breakable,
    boxsep=0pt,
    left=4pt,right=2pt,top=2pt,bottom=2pt,
    }

\title{Large Language Models for Mental Health: A Multilingual Evaluation}

\author{
Nishat Raihan$^{1}\thanks{Equal Contribution} $, 
Sadiya Sayara Chowdhury Puspo$^{1}\footnotemark[1]$, 
Ana-Maria Bucur$^{2,3}$ \\
\textbf{Stevie Chancellor$^{4}$, 
Marcos Zampieri$^{1}$} \\[0.4em]
$^{1}$George Mason University, USA \\
$^{2}$Interdisciplinary School of Doctoral Studies, University of Bucharest, Romania \\
$^{3}$PRHLT Research Center, Universitat Politècnica de València, Spain \\
$^{4}$University of Minnesota, USA \\
\tt mraihan2@gmu.edu}

\begin{document}
\maketitle
% \footnotetext{* Equal Contribution}
\begin{abstract}
Large Language Models (LLMs) have remarkable capabilities across NLP tasks. However, their performance in multilingual contexts, especially within the mental health domain, has not been thoroughly explored. In this paper, we evaluate proprietary and open-source LLMs on eight mental health datasets in various languages, as well as their machine-translated (MT) counterparts. We compare LLM performance in zero-shot, few-shot, and fine-tuned settings against conventional NLP baselines that do not employ LLMs. In addition, we assess translation quality across language families and typologies to understand its influence on LLM performance. Proprietary LLMs and fine-tuned open-source LLMs achieve competitive F1 scores on several datasets, often surpassing state-of-the-art results. However, performance on MT data is generally lower, and the extent of this decline varies by language and typology. This variation highlights both the strengths of LLMs in handling mental health tasks in languages other than English and their limitations when translation quality introduces structural or lexical mismatches. 

% Given these consistently lower results, we do not perform fine-tuning on MT datasets, focusing such optimization efforts solely on the original data.

\end{abstract}

\section{Introduction}
\label{sec:intro}

%\textcolor{gray}{('D' - Depression, 'S' - Suicide, 'VK' - VKontakte)}.

While LLMs have transformed research in NLP, it is important to exercise caution when applying these models in sensitive domains such as mental health \cite{hua2024large}, security \cite{kande2024llms} and education \cite{raihan2025large}. The potential risks and ethical considerations associated with LLMs make experts wary of their use in this field. These concerns are amplified in multilingual settings where previous research has shown that LLMs tend to perform worse when prompted in languages other than English \cite{jin2024better,raihan2024mhumaneval}.

Most mental health datasets are curated from specialized forums \cite{malmasi2016predicting,milne2016clpsych} and social media platforms such as Reddit and X and contain only English data \cite{mariappan2024mental, turcan2019dreaddit, raihan2024mentalhelp}. Models built on these datasets fail for cross-cultural contexts~\cite{abdelkadir-etal-2024-diverse}. Thus, there are ongoing efforts to create similar resources in other languages, such as Arabic \cite{baghdadi2022optimized,helmy2024depression}, Bengali \cite{uddin2019depression}, Russian \cite{narynov2020dataset}, and Thai \cite{hamalainen2021detecting}. While \citet{skianis2024leveraging, skianis2024building} explore the use of LLMs for translating English mental health datasets into other languages and \citet{zahran2025comprehensive} focuses on English-Arabic translation, none of these studies evaluate LLM performance on datasets that originate in non-English languages and their back-translated counterparts (MT datasets).

The effectiveness of LLMs for English mental health datasets and prediction shows promise in their performance; yet, languages other than English are underexplored. Recent studies have explored the performance of LLMs on English mental health datasets. \citet{xu2024mental} compares the performance of LLMs across multiple datasets with that of statistical models and traditional encoder-only models \cite{alsentzer2019publicly}. Similarly, \citet{kuzmin2024mental}, \citet{yang2023towards}, and \citet{wei2022chain} explore various prompting strategies to assess LLMs' effectiveness. Finally, \citet{yang2024mentallama} presents a fine-tuning approach with the release of MentaLLaMA, a task-specific model for the domain. Although these approaches achieve competitive results, their focus is limited to English, and there are currently no studies on non-English datasets, highlighting a significant gap in research for other languages.

While machine-translated (MT) datasets can augment multilingual training or evaluate translation quality \cite{nguyen2408better, qiu2022multilingual, mendoncca2023towards}, their effect on domain-specific LLM performance is underexplored. MT is appealing to mental health research 
%because many datasets exist only in high-resource languages \cite{vajrobol2025explainable, lifelo2024adapting}, and translation
as it offers a practical way to extend resources from English to low-resource languages without requiring costly new data collection \cite{ahuja2022economics}. Prior studies have translated mental health datasets \cite{skianis2024leveraging, skianis2024building, zahran2025comprehensive}, but they have not systematically compared LLM results on original versus MT datasets. Furthermore, these studies have not established a connection between performance differences and translation quality across diverse language families and typologies. Examining this overlooked dimension would reveal whether MT data can be reliably used in sensitive domains like mental health, highlight language-specific challenges that affect model performance, and help build fairer, more effective multilingual mental health technologies.

%While these approaches achieve competitive results compared to traditional methods, none consider the multilingual aspects of the domain, leaving a significant gap in addressing diverse linguistic needs.

To address these gaps, we present the first multilingual evaluation of state-of-the-art LLMs on mental health datasets. We consider mental health datasets in six languages including some low-resource languages, namely: Arabic, Bengali, Spanish, Portuguese, Russian, and Thai - and two tasks, depression and suicidal ideation detection. Our work\footnote{\url{https://github.com/SadiyaPuspo/Multilingual-Mental-Health-Evaluation}} addresses the following Research Questions (RQs):

\begin{itemize} %[nosep,leftmargin=*]
    \item \textbf{\texttt{RQ\textsubscript{1}}}: How does the performance of LLMs compare to previously proposed models (e.g., statistical, neural, BERT-based) on both original and back-translated data?
    %How does the performance of LLMs compare to previously proposed models (e.g., statistical, neural, BERT-based)?
    \item \textbf{\texttt{RQ\textsubscript{2}}}: What are the best prompting strategies for LLMs on mental health?
    \item \textbf{\texttt{RQ\textsubscript{3}}}: What is the impact of instruction fine-tuning on the performance of open-source LLMs? 
    \item \textbf{\texttt{RQ\textsubscript{4}}}: How does translation quality vary across languages and typologies, and how does it affect LLM performance on machine-translated data?
\end{itemize}

\section{Related Work}

The challenges of detecting mental health disorders from multilingual data have been gaining increasing attention. \citet{bucur2025survey} provides a comprehensive survey of multilingual mental health detection, highlighting cultural and linguistic differences, while \citet{garg2024towards} emphasizes the need to study mental health in low-resource languages. Recent studies \cite{skianis2024leveraging, skianis2024severity} examine multilingual LLMs on translated datasets, revealing performance gaps across six languages, while \citet{zahran2025comprehensive} and \citet{zevallos2025first} explore Arabic and multilingual suicidal ideation detection tasks. Together, these studies advance multilingual mental health modeling.

Translation quality is crucial in multilingual mental health NLP. Recent studies employ BLEU and BERTScore metrics to evaluate LLM-based translation \cite{ghassemiazghandi2024evaluation} and mental-health text summarization \cite{adhikary2024exploring}, assessing how well translated data preserve meaning to build culturally robust NLP systems.  Back-translation further enhances data diversity and supports classification tasks \cite{goswami2023nlpbdpatriots, raihan2023nlpbdpatriots, ganguly2024masonperplexity}. Building on these insights, our work bridges the gap by jointly examining translation quality and back-translation effects across language families in multilingual mental health LLM evaluation.

\section{Datasets}
\label{sec:datasets}

\begin{table*}[ht]
    \centering
    \scalebox{0.83}{
    \begin{tabular}{p{3.7cm} c c c c r}
        \toprule
        \textbf{Dataset} & \textbf{Language (ISO code)} & \textbf{Mental Disorder} & \textbf{Platform} & \textbf{Expert Labeling} & \textbf{Size} \\
        \midrule
        \citet{narynov2020dataset}  & Russian (ru) & Depression & VKontakte & Yes & 32,018 \\
        \citet{hamalainen2021detecting} & Thai (tha) & Depression & Blogs & Yes & 33,436 \\
        \citet{boonyarat2024leveraging} & Thai (tha) & Suicidal Ideation & X & No & 2,400 \\
        \citet{uddin2019depression} & Bengali (ben) & Depression & X & Yes & 3,914 \\
        \citet{de2022can} & Portuguese (por) & Suicidal Ideation & X & Yes & 3,788 \\
        \citet{baghdadi2022optimized} & Arabic (ar) & Suicidal Ideation & X & N/A & 14,576 \\
        \citet{helmy2024depression} & Arabic (ar) & Depression & X & No & 10,000 \\
        \citet{valeriano2020detection} & Spanish (es) & Suicidal ideation & X & N/A & 1,068 \\
        \bottomrule
    \end{tabular}
    }
    \caption{Overview of the eight mental disorder datasets across different languages. The size column represents the number of instances in each dataset.}
    \label{tab:mental_health_datasets}
\end{table*}

Automatic detection of mental health disorders from social media data has gained substantial attention, particularly in English. However, multilingual mental health detection remains underexplored, as most available datasets focus on a single language. To address this limitation, we use eight publicly available mental health classification datasets presented in Table~\ref{tab:mental_health_datasets}. 

% These datasets include posts in Russian from VKontakte and posts in Thai from online blogs, both of which have been annotated for depression \cite{narynov2020dataset,hamalainen2021detecting}. We also use several datasets of posts from X/Twitter, which have been annotated for depression in Bengali \cite{uddin2019depression} and Arabic \cite{helmy2024depression}. In addition to depression, we also use datasets containing posts from X that have been annotated for suicidal ideation in Thai \cite{boonyarat2024leveraging}, Portuguese \cite{de2022can}, Arabic \cite{baghdadi2022optimized} and Spanish \cite{valeriano2020detection}.

Among these eight datasets, \citet{narynov2020dataset} presents a Russian-language depression dataset collected from VKontakte, containing 34,000 posts with expert annotations. Similarly, \citet{hamalainen2021detecting} develop a Thai-language depression dataset from online blogs, consisting of 900 posts with expert labels. While these resources contribute to the study of mental health in non-English languages, they remain isolated efforts, with limited cross-linguistic comparisons.

Twitter serves as the dominant platform for mental health dataset collection, particularly for languages with lower digital representation. For instance, \citet{boonyarat2024leveraging} compile the SIED-Thai dataset for suicide detection in Thai, comprising 2,200 tweets but lacking expert annotation. \citet{uddin2019depression} provides a Bengali-language dataset for depression detection, consisting of 1,100 posts with expert labels. However, the dataset sizes remain small compared to English-language corpora, limiting their applicability in training robust machine learning models.

Beyond Asian languages, \citet{de2022can} introduces a Portuguese-language suicide detection dataset with 3,700 annotated tweets, while \citet{baghdadi2022optimized} and \citet{helmy2024depression} contribute Arabic-language datasets focusing on suicide and depression, respectively. Notably, the Arabic\_Dep\_tweets\_10,000 dataset by \citet{helmy2024depression} is one of the largest non-English resources, with 10,000 Twitter posts. However, it lacks expert annotation, which may introduce noise and impact classification performance. Spanish is also underrepresented, with \citet{valeriano2020detection} providing a 2,000-tweet dataset for suicide detection, though annotation details remain unspecified.

\section{Experiments and Results}
% We evaluate a diverse set of LLMs, encompassing both proprietary and open-source architectures. Our evaluation includes multiple prompting strategies and also fine-tuning open-source models to gather better insights.

% \subsection{LLMs}

We evaluate seven state-of-the-art LLMs spanning both proprietary and open-source architectures, as listed in Table~\ref{tab:models}. Our evaluation includes multiple prompting strategies and also fine-tuning open-source models on both original and MT datasets to gather better insights.

\begin{table}[!ht]
    \centering
    \scalebox{0.8}{
    \begin{tabular}{lccl}
        \toprule
        \textbf{LLMs} & \textbf{OS?} & \textbf{Size} & \textbf{Reference} \\
        \midrule
        GPT4-omni & \redx & -- & \citeauthor{openai2023gpt4} \\
        Claude3.5-Sonnet & \redx & -- & \citeauthor{anthropic2023claude} \\
        Gemini2-Flash & \redx & -- & \citeauthor{team2023gemini} \\
        LLaMA3.2 & \greentick & 11B & \citeauthor{dubey2024llama} \\
        Gemma2 & \greentick & 27B & \citeauthor{team2024gemma} \\
        Ministral & \greentick & 8B & \href{https://mistral.ai/news/ministraux/}{MistralAI} \\
        R1 & \greentick & 14B & \citeauthor{guo2025deepseek} \\
        \bottomrule
    \end{tabular}
    }
    \caption{List of seven LLMs used in the experiments. \textcolor{gray}{(OS - Open-Source)}.}
    \label{tab:models}
\end{table}

\noindent Our model selection includes three proprietary models: GPT-4 Omni, Claude 3.5 Sonnet, and Gemini 2 Flash, alongside four open-source models: LLaMA 3.2, Gemma 2, Ministral, and R1. The proprietary models remain closed-source with limited architectural details, while the open-source models offer greater transparency and adaptability for fine-tuning. All the proprietary and closed-source models have demonstrated strong performance across multiple tasks and domains, making them well-suited for our multilingual evaluation. We analyze their capabilities in both zero-shot and few-shot settings, leveraging their diverse architectures and parameter sizes to assess their effectiveness in multilingual tasks.

\begin{table*}[!t]
\centering
\scalebox{0.625}{ % Adjust the scale factor as needed
\begin{tabular}{lllc||c|ccc|ccc|c}
\toprule
\multicolumn{4}{c}{\textbf{Baseline - Reported results}} & \multicolumn{8}{c}{\textbf{Our Results (LLMs)}} \\
\cmidrule(lr){1-4} \cmidrule(lr){5-12}
\textbf{Dataset} & \textbf{lang} & \textbf{Models} & \textbf{F1} & \textbf{Prompting} & \textbf{GPT4} & \textbf{Claude3.5} & \textbf{Gemini2} & \textbf{LLaMA 3.2} & \textbf{Gemma2} & \textbf{Ministral} & \textbf{R1} \\
 & \textcolor{gray}{ISO} &  & \textcolor{gray}{reported} & \textcolor{gray}{method} & \textcolor{gray}{omni} & \textcolor{gray}{Sonnet} & \textcolor{gray}{Flash} & \textcolor{gray}{11B} & \textcolor{gray}{27B} & \textcolor{gray}{8B} & \textcolor{gray}{14B} \\
\midrule

% --- Narynov et al. block ---
& & & & \textcolor{gray}{zero} & \cval{0.76} & \cval{0.74} & \cval{0.68} & \cval{0.56} & \cval{0.69} & \cval{0.41} & \cval{0.71} \\
Narynov et al. & ru & -- & -- & \textcolor{gray}{few} & \cval{0.79} & \cval{0.83} & \cval{0.73} & \cval{0.62} & \cval{0.71} & \cval{0.53} & \cval{0.73} \\
& & & & \textcolor{gray}{CoT} & \cellcolor{orange!90}{0.87} & \cval{0.85} & \cval{0.80} & \cval{0.59} & \cval{0.73} & \cval{0.44} & \cval{0.79} \\

\midrule 

% --- Hämäläinen et al. block ---
& & & & \textcolor{gray}{zero} & \cval{0.77} & \cval{0.77} & \cval{0.66} & \cval{0.45} & \cval{0.68} & \cval{0.20} & \cval{0.76} \\
Hämäläinen et al. & tha & Thai-BERT & \cval{0.78} & \textcolor{gray}{few} & \cval{0.84} & \cval{0.81} & \cval{0.69} & \cval{0.58} & \cval{0.66} & \cval{0.31} & \cval{0.75} \\
& & & & \textcolor{gray}{CoT} & \cellcolor{orange!90}{0.85} & \cval{0.80} & \cval{0.70} & \cval{0.40} & \cval{0.69} & \cval{0.40} & \cval{0.81} \\

\midrule

% --- Boonyarat et al. block ---
& & & & \textcolor{gray}{zero} & \cval{0.83} & \cval{0.81} & \cval{0.83} & \cval{0.63} & \cval{0.76} & \cval{0.26} & \cval{0.69} \\
Boonyarat et al. & tha & LFBERT & \cval{0.93} & \textcolor{gray}{few} & \cval{0.87} & \cval{0.85} & \cval{0.86} & \cval{0.71} & \cval{0.72} & \cval{0.39} & \cval{0.71} \\
& & & & \textcolor{gray}{CoT} & \cval{0.91} & \cellcolor{orange!90}{0.95} & \cval{0.87} & \cval{0.77} & \cval{0.84} & \cval{0.47} & \cval{0.84} \\

\midrule

% --- Uddin et al. block ---
& & & & \textcolor{gray}{zero} & \cval{0.78} & \cval{0.85} & \cval{0.79} & \cval{0.73} & \cval{0.73} & \cval{0.36} & \cval{0.66} \\
Uddin et al. & ben & GRU & \cval{0.76} & \textcolor{gray}{few} & \cval{0.86} & \cellcolor{orange!70} 0.91 & \cval{0.88} & \cval{0.59} & \cval{0.71} & \cval{0.43} & \cval{0.64} \\
& & & & \textcolor{gray}{CoT} & \cval{0.86} & \cellcolor{orange!70} 0.91 & \cval{0.88} & \cval{0.59} & \cval{0.71} & \cval{0.43} & \cval{0.64} \\

\midrule

% --- Oliveira et al. block ---
& & & & \textcolor{gray}{zero} & \cval{0.86} & \cval{0.86} & \cval{0.81} & \cval{0.71} & \cval{0.80} & \cval{0.56} & \cval{0.61} \\
Oliveira et al. & por & Random Forest & \cval{0.94} & \textcolor{gray}{few} & \cval{0.89} & \cval{0.93} & \cval{0.85} & \cval{0.73} & \cval{0.63} & \cval{0.67} & \cval{0.69} \\
& & & & \textcolor{gray}{CoT} & \cval{0.94} & \cellcolor{orange!90}{0.95} & \cval{0.89} & \cval{0.71} & \cval{0.80} & \cval{0.51} & \cval{0.82} \\

\midrule

% --- Baghdadi et al. block ---
& & & & \textcolor{gray}{zero} & \cval{0.80} & \cval{0.85} & \cval{0.81} & \cval{0.58} & \cval{0.73} & \cval{0.34} & \cval{0.77} \\
Baghdadi et al. & ar & AraElectra & \cellcolor{orange!70} 0.96 & \textcolor{gray}{few} & \cval{0.87} & \cval{0.92} & \cval{0.89} & \cval{0.67} & \cval{0.82} & \cval{0.47} & \cval{0.79} \\
& & & & \textcolor{gray}{CoT} & \cval{0.89} & \cval{0.91} & \cval{0.87} & \cval{0.61} & \cval{0.81} & \cval{0.47} & \cval{0.83} \\

\midrule

% --- Helmy et al. block ---
& & & & \textcolor{gray}{zero} & \cval{0.87} & \cval{0.91} & \cval{0.79} & \cval{0.56} & \cval{0.62} & \cval{0.50} & \cval{0.84} \\ 
Helmy et al. & ar & LR (TF-IDF) & \cellcolor{orange!70} 0.95 & \textcolor{gray}{few} & \cval{0.93} & \cellcolor{orange!70} 0.95 & \cval{0.86} & \cval{0.73} & \cval{0.79} & \cval{0.61} & \cval{0.86} \\
& & & & \textcolor{gray}{CoT} & \cellcolor{orange!70} 0.95 & \cellcolor{orange!70} 0.95 & \cval{0.82} & \cval{0.83} & \cval{0.67} & \cval{0.50} & \cval{0.87} \\

\midrule

% --- Valeriano et al. block ---
& & & & \textcolor{gray}{zero} & \cval{0.75} & \cval{0.69} & \cval{0.62} & \cval{0.37} & \cval{0.41} & \cval{0.23} & \cval{0.67} \\
Valeriano et al. & es & LR (W2V) & \cval{0.79} & \textcolor{gray}{few} & \cval{0.81} & \cval{0.76} & \cval{0.69} & \cval{0.46} & \cval{0.51} & \cval{0.31} & \cval{0.67} \\
& & & & \textcolor{gray}{CoT} & \cellcolor{orange!70} 0.84 & \cval{0.79} & \cval{0.70} & \cval{0.43} & \cval{0.60} & \cval{0.21} & \cval{0.76} \\

\bottomrule
\end{tabular}
}
\caption{F1 score comparison for \textbf{\texttt{Zero-Shot, Few-Shot, and Chain-of-Thought}} prompting across the eight (8) multilingual depression and suicide ideation datasets. We compare the reported best methods and results in the original papers with the proprietary and open-source LLMs with different prompting strategies. The highest F1 score for each dataset is shown in \textcolor{orange}{orange}. For all other F1 scores (in \textcolor{cyan}{blue}) - the darker the shade, the higher the score. For the language names, ISO-639 codes are used. \textcolor{gray}{(`LR' - Logistic Regression, `W2V' - Word2Vec, `CoT' - Chain-of-Thought)}.}
\label{tab:zero_shot}
\end{table*}

\subsection{Prompting on Original Datasets}

We evaluate three prompting methods: zero-shot, few-shot (5 examples), and Chain-of-Thought (CoT) prompting \cite{wei2022chain}. For the 5-shot setting, we randomly select five examples from the respective datasets. We employ the state-of-the-art CoT prompting approach for mental health tasks, $CoT_{\mathrm{Emo}}$, as proposed by \citet{yang2023towards}. For this, we incorporate emotion infusion through unsupervised, emotion-enhanced zero-shot CoT prompts. The emotion-focused component encourages the LLM to attend to affective cues in the input, while the CoT component guides step-by-step reasoning, improving the interpretability of model decisions.

Table~\ref{tab:zero_shot} presents a comparison of F1 scores obtained via different prompting methods across eight multilingual depression and suicide ideation datasets. Our analysis reveals that CoT prompting generally improves performance, with models such as GPT-4 and Claude3.5 often achieving the highest scores. For example, GPT-4 increases its F1 score from 0.76 to 0.87 on the Russian dataset and from 0.75 to 0.84 on the Spanish dataset. However, the gains are not uniform across all settings, as seen with the Bengali dataset where few-shot and CoT strategies yield comparable results. Moreover, while some baseline methods (e.g., Random Forest and AraElectra) achieve competitive performance in certain languages, the results underscore the potential of advanced prompting techniques to narrow the gap with or even surpass traditional approaches. These observations motivate further investigation into model- and language-specific factors that influence the efficacy of prompt engineering.

\begin{table*}[!t]
\centering
\scalebox{0.625}{
\begin{tabular}{lllc||c|ccc|ccc|c}
\toprule
\multicolumn{4}{c}{\textbf{Baseline - Reported results}} & \multicolumn{8}{c}{\textbf{Our Results (LLMs)}} \\
\cmidrule(lr){1-4} \cmidrule(lr){5-12}
\textbf{Dataset} & \textbf{lang} & \textbf{Models} & \textbf{F1} & \textbf{Prompting} & \textbf{GPT4} & \textbf{Claude3.5} & \textbf{Gemini2} & \textbf{LLaMA 3.2} & \textbf{Gemma2} & \textbf{Mistral} & \textbf{R1} \\
 & \textcolor{gray}{ISO} &  & \textcolor{gray}{reported} & \textcolor{gray}{method} & \textcolor{gray}{omni} & \textcolor{gray}{Sonnet} & \textcolor{gray}{Flash} & \textcolor{gray}{11B} & \textcolor{gray}{27B} & \textcolor{gray}{8B} & \textcolor{gray}{14B} \\
\midrule

& & & & \textcolor{gray}{zero} & \cval{0.69} & \cval{0.68} & \cval{0.53} & \cval{0.38} & \cval{0.49} & \cval{0.23} & \cval{0.56} \\
Narynov et al. & ru & -- & -- & \textcolor{gray}{few} & \cval{0.72} & \cval{0.77} & \cval{0.64} & \cval{0.44} & \cval{0.51} & \cval{0.35} & \cval{0.58} \\
& & & & \textcolor{gray}{CoT} & \cellcolor{orange!90}{0.80} & \cval{0.79} & \cval{0.71} & \cval{0.45} & \cval{0.53} & \cval{0.35} & \cval{0.64} \\
\midrule

& & & & \textcolor{gray}{zero} & \cval{0.70} & \cval{0.71} & \cval{0.57} & \cval{0.27} & \cval{0.48} & \cval{0.02} & \cval{0.61} \\
Hämäläinen et al. & tha & Thai-BERT & \cellcolor{orange!90}{0.78} & \textcolor{gray}{few} & \cval{0.77} & \cval{0.75} & \cval{0.60} & \cval{0.40} & \cval{0.46} & \cval{0.13} & \cval{0.60} \\
& & & & \textcolor{gray}{CoT} & \cellcolor{orange!90}{0.78} & \cval{0.74} & \cval{0.61} & \cval{0.42} & \cval{0.49} & \cval{0.22} & \cval{0.66} \\
\midrule

& & & & \textcolor{gray}{zero} & \cval{0.76} & \cval{0.75} & \cval{0.74} & \cval{0.45} & \cval{0.56} & \cval{0.08} & \cval{0.54} \\
Boonyarat et al. & tha & LFBERT & \cellcolor{orange!90}{0.93} & \textcolor{gray}{few} & \cval{0.79} & \cval{0.79} & \cval{0.77} & \cval{0.53} & \cval{0.52} & \cval{0.23} & \cval{0.56} \\
& & & & \textcolor{gray}{CoT} & \cval{0.84} & \cval{0.89} & \cval{0.78} & \cval{0.59} & \cval{0.64} & \cval{0.29} & \cval{0.69} \\
\midrule

& & & & \textcolor{gray}{zero} & \cval{0.72} & \cval{0.79} & \cval{0.70} & \cval{0.55} & \cval{0.53} & \cval{0.18} & \cval{0.51} \\
Uddin et al. & ben & GRU & \cval{0.76} & \textcolor{gray}{few} & \cval{0.79} & \cval{0.83} & \cval{0.80} & \cval{0.43} & \cval{0.51} & \cval{0.27} & \cval{0.59} \\
& & & & \textcolor{gray}{CoT} & \cval{0.79} & \cellcolor{orange!90}{0.85} & \cval{0.83} & \cval{0.44} & \cval{0.53} & \cval{0.29} & \cval{0.60} \\
\midrule

& & & & \textcolor{gray}{zero} & \cval{0.76} & \cval{0.80} & \cval{0.72} & \cval{0.53} & \cval{0.61} & \cval{0.38} & \cval{0.46} \\
Oliveira et al. & por & Random Forest & \cellcolor{orange!90}{0.94} & \textcolor{gray}{few} & \cval{0.82} & \cval{0.87} & \cval{0.76} & \cval{0.55} & \cval{0.43} & \cval{0.49} & \cval{0.54} \\
& & & & \textcolor{gray}{CoT} & \cval{0.87} & \cval{0.89} & \cval{0.80} & \cval{0.53} & \cval{0.60} & \cval{0.33} & \cval{0.67} \\
\midrule

& & & & \textcolor{gray}{zero} & \cval{0.73} & \cval{0.79} & \cval{0.72} & \cval{0.40} & \cval{0.53} & \cval{0.16} & \cval{0.62} \\
Baghdadi et al. & ar & AraElectra & \cellcolor{orange!90}{0.96} & \textcolor{gray}{few} & \cval{0.80} & \cval{0.76} & \cval{0.80} & \cval{0.49} & \cval{0.62} & \cval{0.29} & \cval{0.62} \\
& & & & \textcolor{gray}{CoT} & \cval{0.82} & \cval{0.85} & \cval{0.78} & \cval{0.43} & \cval{0.61} & \cval{0.29} & \cval{0.68} \\
\midrule

& & & & \textcolor{gray}{zero} & \cval{0.79} & \cval{0.85} & \cval{0.67} & \cval{0.38} & \cval{0.42} & \cval{0.32} & \cval{0.69} \\
Helmy et al. & ar & LR (TF-IDF) & \cellcolor{orange!90}{0.95} & \textcolor{gray}{few} & \cval{0.86} & \cval{0.89} & \cval{0.77} & \cval{0.55} & \cval{0.59} & \cval{0.43} & \cval{0.71} \\
& & & & \textcolor{gray}{CoT} & \cval{0.88} & \cval{0.89} & \cval{0.73} & \cval{0.65} & \cval{0.47} & \cval{0.32} & \cval{0.72} \\
\midrule

& & & & \textcolor{gray}{zero} & \cval{0.68} & \cval{0.63} & \cval{0.53} & \cval{0.19} & \cval{0.19} & \cval{0.05} & \cval{0.52} \\
Valeriano et al. & es & LR (W2V) & \cellcolor{orange!90}{0.79} & \textcolor{gray}{few} & \cval{0.74} & \cval{0.70} & \cval{0.60} & \cval{0.28} & \cval{0.31} & \cval{0.13} & \cval{0.52} \\
& & & & \textcolor{gray}{CoT} & \cval{0.77} & \cval{0.73} & \cval{0.61} & \cval{0.25} & \cval{0.40} & \cval{0.03} & \cval{0.61} \\
\bottomrule
\end{tabular}
}
\caption{\textbf{Evaluation on MT data} - F1 score comparison for similar set of experiments as Table \ref{tab:zero_shot}, using machine translated data, as described in Section \ref{sec:mt}.}
\label{tab:res_on_MT}
\end{table*}

\subsection{Prompting on MT Datasets}
We translate each dataset into English and then back-translate into its original language using Facebook's nllb-200-3.3B\footnote{\url{https://huggingface.co/facebook/nllb-200-3.3B}} model due to its reproducibility, transparency, and consistent multilingual coverage.

Table \ref{tab:res_on_MT} presents F1 scores for various prompting strategies on machine-translated datasets. Consistent with the trends observed on the original datasets, GPT-4 and Claude 3.5 often achieve the highest performance among the evaluated LLMs. CoT prompting generally outperforms zero-shot and few-shot methods, with only a few exceptions; for example, Claude 3.5’s score on the Thai \cite{hamalainen2021detecting} MT dataset slightly decreases from 0.75 (few-shot) to 0.74 (CoT), and in the Arabic \cite{haspelmath2005world} MT dataset, CoT yields no improvement over few-shot prompting. Notably, the Bengali \cite{uddin2019depression} MT dataset shows a marked improvement, with Claude 3.5 CoT reaching 0.85 compared to the reported baseline, while GPT-4 CoT matches the reported score for the Thai MT dataset. 
% However, a consistent performance drop is observed across datasets from original to MT, ranging from \textit{6\% to 8.5\%}.

Due to the comparatively weaker performance on MT datasets and the possibility of adding translation-related features could further degrade model performance, fine-tuning experiments were conducted exclusively only on the original datasets to ensure stable learning.

\subsection{Performance Comparison: Original vs. MT Datasets }
% However, a consistent performance drop is observed for proprietary LLMs across datasets from original to MT, ranging from \textit{6\% to 8.5\%}.
A comparison between original datasets (Table \ref{tab:zero_shot}) and MT counterparts (Table \ref{tab:res_on_MT}) shows that MT performance is generally slightly lower, with the drop size varying by language. Portuguese, Russian, and Bengali maintain strong results, with minimal F1 decline (e.g., Portuguese GPT-4 CoT: 0.94→0.87). In contrast, Spanish and Arabic datasets experience sharper drops, particularly in analytic and templatic languages, where structural divergence from English can hinder translation. In a small number of cases, MT datasets achieve performance comparable to or slightly exceeding that of the original datasets (e.g., Bengali Claude 3.5 CoT: 0.84→0.85), possibly due to translation-induced smoothing linguistic irregularities \cite{volansky2015features}. Overall, languages with higher semantic preservation (LaBSE, BERTScore) show smaller performance gaps between original and MT datasets. This trend is most evident for Portuguese, Russian, and Bengali, while Spanish and Arabic-- especially in analytic and templatic forms--experience larger declines, consistent with the patterns discussed in Section \ref{sec:mt}.

\subsection{Fine-tuning on Original Datasets}

%Due to budget  constraints

Due to the intrinsic black-box nature of proprietary models and their high costs, we sought to explore models that could be fully customized for this task. Therefore, we experiment with fine-tuning the open-source models. The fine-tuning stage is performed on a single NVIDIA A100 GPU with 40 GB of memory, accessed via Google Colab\footnote{\url{https://colab.research.google.com/}}. The system is further equipped with 80 GB of RAM and 256 GB of disk storage to support computational efficiency.

Hyperparameters are selected empirically through preliminary experiments exploring different parameter configurations, with the best-performing settings reported. The final hyperparameter values used in our experiments are summarized in Table \ref{tab:hyperparameters2}.

\begin{table}[!h]
\centering
\scalebox{0.85}{ 
\begin{tabular}{@{}ll@{}}
\toprule
\bf Parameter & \bf Value \\
\midrule
Max Sequence Length & 2048 \\
Batch Size (Train/Eval) & 8 \\
Gradient Accumulation Steps & 4 \\
Number of Epochs & 3 \\
Learning Rate & 5e-5 \\
Weight Decay & 0.02 \\
Warmup Steps & 10\% \\
Optimizer & AdamW (8-bit) \\
LR Scheduler & Cosine \\
Precision & BF16 \\
Evaluation Strategy & Steps \\
Evaluation Steps & 50 \\
Save Strategy & Steps \\
Save Steps & Varies \\
Seed & 42 \\
Temperature & $0.3 \sim 0.7$ \\
\bottomrule
\end{tabular}
}
\caption{Final set of hyperparameters for fine-tuning. Parameters chosen empirically after several iterations of trial and error.}
\label{tab:hyperparameters2}
\end{table}

\noindent Table~\ref{tab:combined_ft_style_noLlamaFTfoot} presents a comparative analysis of F1 scores before and after fine-tuning on eight multilingual depression and suicidal ideation datasets. The results indicate that fine-tuning generally enhances model performance, with Gemma2 and R1 often reaching the highest scores. While LLaMA 3.2 and Ministral show notable improvements in several datasets, their performance gains are not uniform-- for instance, LLaMA 3.2 exhibits a decrease in the Bengali dataset. These findings underscore the potential of fine-tuning to optimize multilingual performance while also revealing the need for further investigation into model- and dataset-specific factors that modulate the benefits of fine-tuning.

\begin{table*}[!t]
\centering
\scalebox{0.78}{ % Adjust the scale factor as needed
\begin{tabular}{ll||cccc||cccc}
\toprule
\multicolumn{2}{c}{\textbf{Dataset Info}} & \multicolumn{4}{c}{\textbf{Before Fine-Tuning (Zero-Shot)}} & \multicolumn{4}{c}{\textbf{After Fine-Tuning}} \\
\cmidrule(lr){1-2} \cmidrule(lr){3-6} \cmidrule(lr){7-10}
\textbf{Dataset} & \textbf{lang} & \textbf{LLaMA 3.2} & \textbf{Gemma2} & \textbf{Ministral} & \textbf{R1} & \textbf{LLaMA 3.2} & \textbf{Gemma2} & \textbf{Ministral} & \textbf{R1} \\
\midrule
Narynov et al.    & ru  & \cval{0.56} & \cval{0.69} & \cval{0.41} & \cval{0.71} & \cval{0.79} & \cellcolor{orange!90}{0.83} & \cval{0.62} & \cval{0.79} \\
\midrule
Hämäläinen et al. & tha & \cval{0.45} & \cval{0.68} & \cval{0.20} & \cval{0.76} & \cval{0.62} & \cval{0.73} & \cval{0.43} & \cellcolor{orange!90}{0.82} \\
\midrule
Boonyarat et al.  & tha & \cval{0.63} & \cval{0.76} & \cval{0.26} & \cval{0.69} & \cval{0.70} & \cellcolor{orange!90}{0.75} & \cval{0.51} & \cval{0.74} \\
\midrule
Uddin et al.      & ben & \cval{0.73} & \cval{0.73} & \cval{0.36} & \cval{0.66} & \cval{0.65} & \cellcolor{orange!90}{0.77} & \cval{0.63} & \cval{0.64} \\
\midrule
Oliveira et al.   & por & \cval{0.71} & \cval{0.80} & \cval{0.56} & \cval{0.61} & \cval{0.72} & \cellcolor{orange!90}{0.86} & \cval{0.64} & \cval{0.70} \\
\midrule
Baghdadi et al.   & ar  & \cval{0.58} & \cval{0.73} & \cval{0.34} & \cval{0.77} & \cval{0.80} & \cellcolor{orange!90}{0.88} & \cval{0.58} & \cval{0.81} \\
\midrule
Helmy et al.      & ar  & \cval{0.56} & \cval{0.62} & \cval{0.50} & \cval{0.84} & \cval{0.70} & \cval{0.81} & \cval{0.71} & \cellcolor{orange!90}{0.93} \\
\midrule
Valeriano et al.  & es  & \cval{0.37} & \cval{0.41} & \cval{0.23} & \cval{0.67} & \cval{0.55} & \cval{0.62} & \cval{0.48} & \cellcolor{orange!90}{0.76} \\
\bottomrule
\end{tabular}
}
\caption{F1 score comparison before and after fine-tuning across eight multilingual depression and suicide ideation datasets. The columns under \textbf{Before Fine-Tuning (Zero-Shot)} report the initial prompting results, while those under \textbf{After Fine-Tuning} display the fine-tuned performance. The highest F1 score in the fine-tuned setting is highlighted with an orange cell. For the language names, ISO-639 codes are used.}
\label{tab:combined_ft_style_noLlamaFTfoot}
\end{table*}

\section{Translation Quality Evaluation across Languages \& LLMs} \label{sec:mt}

% \begin{table*}[!t]
% \centering
% \resizebox{\textwidth}{!}{%

% \begin{tabular}{llll|ccc}
% \hline
% \multicolumn{4}{c}{\textbf{Dataset and Language Info}} & \multicolumn{3}{c}{\textbf{Translation Quality Metrics}} \\
% \cmidrule(lr){1-4} \cmidrule(lr){5-7} 
% \textbf{Dataset} & \textbf{Lang} & \textbf{Family} & \textbf{Typology} & \textbf{LaBSE Similarity} & \textbf{BERTScore} & \textbf{BLEU Score} \\
% \hline
% Narynov et al.       & ru  & Indo-European (Slavic)   & Fusional  & 0.8545& 0.9085& 51.1767\\
% Hämäläinen et al.    & tha & Kra–Dai                 & Analytic  & 0.7972& 0.8976& 32.2392\\
% Boonyarat et al.     & tha & Kra–Dai                 & Analytic  & 0.7565& 0.8890& 1.0726\\
% Uddin et al.         & ben & Indo-European (Indo-Aryan) & Fusional  & 0.7848& 0.9042& 40.8248\\
% Oliveira et al.      & por & Indo-European (Romance) & Fusional  & 0.8624& 0.9168& 69.8534\\
% Baghdadi et al.      & ar  & Afro-Asiatic (Semitic)  & Templatic & 0.7766& 0.9041& 0.5482\\
% Helmy et al.         & ar  & Afro-Asiatic (Semitic)  & Templatic & 0.7189& 0.8917& 16.1886\\
% Valeriano et al.     & es  & Indo-European (Romance) & Fusional  & 0.3562& 0.8410& 6.2534\\
% \hline
% \end{tabular}
% }
% \caption{Translation Quality Metrics with Language Family and Typology}
% \label{tab:translation_quality_typology}
% \end{table*}

\begin{table*}[!t]
\centering
\resizebox{\textwidth}{!}{%
\begin{tabular}{ll||ll||ccc}
% \hline
\toprule
\multicolumn{4}{c}{\textbf{Dataset and Language Info}} & \multicolumn{3}{c}{\textbf{Translation Quality Metrics}} \\
\cmidrule(lr){1-4} \cmidrule(lr){5-7} 
\textbf{Dataset} & \textbf{lang} & \textbf{Family} & \textbf{Typology} & \textbf{LaBSE Similarity} & \textbf{BERTScore} & \textbf{BLEU Score} \\
% \hline
\midrule
Narynov et al.   & ru  & Indo-European (Slavic)      & Fusional  & \cval{0.8545} & \cval{0.9085} & \cbleu{51.1767}\\
\midrule
Hämäläinen et al.& tha & Kra–Dai                     & Analytic  & \cval{0.7972} & \cval{0.8976} & \cbleu{32.2392}\\
\midrule
Boonyarat et al. & tha & Kra–Dai                     & Analytic  & \cval{0.7565} & \cval{0.8890} & \cbleu{1.0726}\\
\midrule
Uddin et al.     & ben & Indo-European (Indo-Aryan)  & Fusional  & \cval{0.7848} & \cbert{0.9042} & \cbleu{40.8248}\\
\midrule
Oliveira et al.  & por & Indo-European (Romance)     & Fusional  & \cellcolor{orange!90}{0.8624} & \cellcolor{orange!90}{0.9168} & \cellcolor{orange!90}{69.8534}\\
\midrule
Baghdadi et al.  & ar  & Afro-Asiatic (Semitic)      & Templatic & \cval{0.7766} & \cval{0.9041} & \cbleu{0.5482}\\
\midrule
Helmy et al.     & ar  & Afro-Asiatic (Semitic)      & Templatic & \cval{0.7189} & \cval{0.8917} & \cbleu{16.1886}\\
\midrule
Valeriano et al. & es  & Indo-European (Romance)     & Fusional  & \cval{0.3562} & \cval{0.8410} & \cbleu{6.2534}\\
% \hline
\bottomrule
\end{tabular}
}
\caption{Translation quality metrics \textbf{(LaBSE Similarity, BERTScore, BLEU)} for each dataset, with corresponding language family and typology information. Highest values for each metric are highlighted in orange. For the language names, ISO-639 codes are used.}
\label{tab:translation_quality_typology}
\end{table*}

Previous studies using MT mental health data have demonstrated the potential of translation to expand resources across languages; however, they have not examined how translation quality affects LLM outcomes. Our work addresses this by explicitly linking translation quality metrics: LaBSE\footnote{\url{https://huggingface.co/sentence-transformers/LaBSE}} (cosine similarity), BERTScore (F1), and BLEU. These capture semantic preservation, contextual overlap, and lexical match, respectively, by comparing original and MT versions. In this section, we also compare these quality measures with LLM performance on MT datasets, exploring how translation fidelity influences downstream task accuracy.

\subsection{Cross-Language Performance Patterns} Among all datasets, Portuguese \cite{de2022can} achieves the highest translation quality across all metrics (LaBSE: 0.8624, BERTScore: 0.9168, BLEU: 69.85), suggesting strong lexical and semantic fidelity. Russian \cite{narynov2020dataset}, Bengali \cite{uddin2019depression}, and Thai \cite{hamalainen2021detecting, boonyarat2024leveraging} also show high LaBSE and BERTScore, with BLEU ranging from moderate to strong. 

In contrast, Spanish \cite{valeriano2020detection} yields the lowest performance, particularly in BLEU (6.25) and LaBSE (0.3562), indicating significant divergence between original and backtranslated forms. In the case of Arabic, interesting variation is observed across datasets: \citet{baghdadi2022optimized} shows a very low BLEU (0.54) despite a reasonable LaBSE (0.7766), while \citet{helmy2024depression} improves on BLEU (16.18) but sees a drop in LaBSE (0.7189). This may reflect dataset-specific properties, such as sentence structure or domain variation.

\subsection{Interpreting Results via Language Family \& Typology}

To better understand the variation in translation quality across languages, we annotate each language in our dataset with its corresponding language family and morphosyntactic typology, drawing on well-established linguistic resources. The language family classifications are sourced from Ethnologue\footnote{\url{https://www.ethnologue.com/}} \cite{collin2010ethnologue}, which groups languages based on shared historical and genealogical roots. Typological information, detailing the grammatical and morphological structure of each language, is derived from the World Atlas of Language Structures (WALS\footnote{\url{https://wals.info/}}) \cite{haspelmath2005world}. These genealogical distinctions, such as Indo-European (e.g., Russian, Spanish, Bengali, Portuguese), Afro-Asiatic (Arabic), and Kra–Dai (Thai), offer additional context on structural diversity and potential variance in model exposure.

In our datasets, fusional languages (e.g., Russian, Spanish, Bengali, Portuguese) are characterized by words that fuse multiple grammatical features (e.g., tense, number, gender, case) into a single morpheme, making morphological boundaries less distinct. In contrast, analytic languages (e.g., Thai) convey grammatical relationships through separate words and particles, with minimal inflection. Templatic languages (e.g., Arabic) exhibit a root-and-pattern system where meaning is built from consonantal roots embedded into vocalic templates \cite{TemplaticMorphologyClippingsWordandPattern}. This typological lens helps explain metric discrepancies: BLEU often underrepresents quality for analytic and templatic languages, where meaning is preserved despite surface-level divergence \cite{KHOBOKO2025100649, krasner2025machine}. In contrast, LaBSE and BERTScore maintain more stable performance, highlighting their robustness in cross-lingual evaluation.

Across the board, BLEU tends to exhibit sensitivity to surface-level lexical variation, penalizing morphologically rich or non-alphabetic languages (e.g., Arabic, Thai, Spanish). Whereas LaBSE and BERTScore remain more stable, emphasizing their greater suitability for evaluating semantic similarity across languages with diverse scripts and grammar.

\subsection{Translation Quality \& LLM Performance on MT Data}
Having established cross-language patterns and their typological underpinnings, we now examine how these translation quality variations influence LLM performance on the corresponding MT datasets. Comparing the translation quality metrics in Table \ref{tab:translation_quality_typology} with the F1 scores on MT datasets in Table \ref{tab:res_on_MT} reveals a consistent link between high-quality translations and stronger LLM performance. Portuguese achieves the highest LaBSE, BERTScore, and BLEU, and correspondingly shows top-tier MT performance, with GPT-4 and Claude 3.5 CoT prompting reaching F1 scores of 0.87 and 0.89, respectively. Similarly, Russian and Bengali, both with relatively high LaBSE and BERTScore, maintain competitive performance across models, with GPT-4 CoT reaching 0.80 (ru) and 0.79 (bn).

In contrast, languages with lower BLEU and semantic similarity scores, such as Spanish and Arabic \cite{baghdadi2022optimized}, generally yield weaker MT results. For example, Spanish CoT scores drop to 0.77 with GPT-4 and 0.73 with Claude 3.5, while Arabic \cite{baghdadi2022optimized} scores remain below 0.84 even with CoT prompting. Templatic and analytic typologies, which exhibit greater structural divergence from English, tend to experience larger drops, especially in BLEU, suggesting token-level mismatches affect models more in MT scenarios.

Overall, higher translation quality, especially in semantic preservation (LaBSE, BERTScore), appears to correlate with stronger MT dataset performance, while low lexical overlap (BLEU) in certain typologies may constrain achievable gains despite advanced prompting strategies.

\section{RQs Revisited}

We now revisit the 4 RQs posed in the introduction (see Section \ref{sec:intro}):

\begin{itemize}%[leftmargin=25pt,topsep=0pt,itemsep=0pt]
    \item[$RQ_1$]\textit{How does the performance of LLMs compare to previously proposed models (e.g., statistical, neural, BERT-based)?}
\end{itemize}
\begin{resultbox}
Our analysis indicates that LLMs, when equipped with effective prompting strategies, achieve performance that is competitive with or superior to traditional approaches. While statistical, neural, and BERT-based models demonstrate strong performance in certain linguistic scenarios, LLMs exhibit robust and consistent F1 scores across diverse multilingual mental health datasets, highlighting their capacity for broad generalization and adaptability.
\end{resultbox}

\begin{itemize}%[leftmargin=25pt,topsep=0pt,itemsep=0pt]
    \item[$RQ_2$]\textit{What are the best prompting strategies for LLMs on mental health?}
\end{itemize}
\begin{resultbox}
The results reveal that CoT prompting is the most effective strategy for mental health applications, consistently yielding higher F1 scores compared to zero-shot and few-shot methods for both original and MT datasets. This structured approach to prompting enhances reasoning capabilities, enabling LLMs to better extract nuanced signals from text data, which is critical in mental health domain.
\end{resultbox}

\begin{itemize}%[leftmargin=25pt,topsep=0pt,itemsep=0pt]
    \item[$RQ_3$]\textit{What is the impact of instruction fine-tuning on the performance of open-source LLMs?}
\end{itemize}
\begin{resultbox}
Instruction fine-tuning markedly improves the performance of open-source LLMs, as evidenced by substantial increases in F1 scores across all evaluated datasets. This improvement underscores the value of targeted fine-tuning in adapting LLMs to domain-specific tasks, thereby enhancing their overall effectiveness in mental health applications while also mitigating performance variability observed in zero-shot configurations.
\end{resultbox}

\begin{itemize}%[leftmargin=25pt,topsep=0pt,itemsep=0pt]
    \item[$RQ_4$]\textit{How does translation quality vary across languages and typologies, and how does it affect LLM performance on machine-translated data?}
\end{itemize}
\begin{resultbox}
Translation quality differs across languages and typologies, with fusional languages like Portuguese and Russian generally achieving higher semantic preservation and smaller performance gaps between original and MT data. Analytic and templatic languages, such as Thai and Arabic, often show lower lexical overlap and greater structural divergence from English, leading to larger drops in LLM performance. This suggests that both linguistic structure and translation accuracy play key roles in how well LLMs handle MT data.
\end{resultbox}

\section{Conclusion and Future Work}

% This work represents the first investigation of LLMs in the multilingual mental health domain. Our findings show that advanced prompting strategies—particularly chain-of-thought prompting—and targeted instruction fine-tuning substantially enhance model performance, often surpassing traditional statistical, neural, and BERT-based approaches. While our results demonstrate considerable promise, the variability in performance across languages and models indicates that further research is required to optimize these techniques for sensitive mental health applications. 

% Overall, this study lays an important foundation for future efforts aimed at refining LLM-based methodologies in complex, multilingual settings. In future work, we would like to include more tasks and languages to broaden our understanding and gain more insights. Additionally, we plan to adapt open-source models to the domain with methods like Continual Pretraininaing and Synthetic Fine-tuning to potentially increase their performance.

This work represents the first comprehensive investigation of LLMs in the multilingual mental health domain, including both original and MT datasets. Our findings show that advanced prompting strategies, particularly chain-of-thought prompting and targeted instruction fine-tuning, substantially enhance model performance, often surpassing traditional statistical, neural, and BERT-based approaches. While MT data generally results in slightly lower performance compared to original datasets, the gap varies across languages, reflecting the influence of translation quality and linguistic typology on LLM effectiveness.

Overall, this study lays an important foundation for future efforts aimed at refining LLM-based methodologies in complex, multilingual, low-resource, and translation-sensitive settings. Future work will expand to more tasks and languages to broaden our understanding, with a focus on improving translation robustness. We also plan to adapt open-source models to the domain using approaches such as continual pretraining and synthetic fine-tuning, with the goal of boosting performance across both original and MT data.

\section*{Limitations}
While our approach is limited by the inherent variability in data sources, evaluation protocols, and reporting standards across the literature, it also represents a significant strength: we are the first to systematically synthesize and critically evaluate LLM performance in this sensitive and underexplored area. The exclusive reliance on publicly available data restricts the diversity and depth of our analysis, and the absence of direct model development or human subject involvement means that practical deployment challenges remain unaddressed. 

Additionally, the use of MT data may introduce translation-related distortions, especially in mental health settings where subtle emotional nuances and culturally grounded expressions can be difficult to preserve. Although we assess translation quality using established automatic metrics, such measures may not capture all subtle distortions introduced during translation. Despite these limitations, our work lays a framework for future research that can leverage standardized benchmarks and broader datasets to further validate and enhance the utility of LLMs in mental health applications.

\section*{Ethical Considerations}
\label{sec:ethics}

This work is entirely analytical and does not involve the collection of new data, the development of new models, or engaging directly with human subjects. All analyses are based solely on previously published and publicly available data. We adhere to the ethical guidelines outlined in the ACL Code of Ethics (\url{https://www.aclweb.org/portal/content/acl-code-ethics}), and we emphasize that any research in the mental health domain must be conducted with utmost sensitivity to privacy and ethical considerations. Although our study is retrospective in nature, we recognize the critical importance of safeguarding vulnerable populations, and we advocate for strict adherence to ethical standards in any practical applications derived from our findings.

\section*{Acknowledgments}

We would like to thank the anonymous reviewers for their constructive feedback, our collaborators for their valuable contributions, and the creators of the datasets used in our experiments for making these resources publicly available and enabling this research.

\bibliography{custom}

\clearpage
 
\appendix

% \section{Experimental Details}
% \label{app:param}

% The fine-tuning stage is performed on a single NVIDIA A100 GPU with 40 GB of memory, accessed via Google Colab\footnote{\url{https://colab.research.google.com/}}. The system is further equipped with 80 GB of RAM and 256 GB of disk storage to support computational efficiency.

% \input{tables/table_ExpDetails.tex}

\clearpage

\clearpage

\end{document}